\documentclass{article}
\usepackage{spconf,amsmath,graphicx}
\usepackage{hyperref}

\title{Caption-Driven Explainability: Probing CNNs for Bias via CLIP}
\usepackage{ifthen}
\newboolean{anonsubmission}
\setboolean{anonsubmission}{false} 

\ifthenelse{\boolean{anonsubmission}}{
    \name{\vspace{1.2em}}
    \address{\vspace{1em}}
}{
    \name{Patrick Koller$^{1}$ \qquad Amil V. Dravid$^{2}$ \qquad Guido M. Schuster$^{3}$ \qquad Aggelos K. Katsaggelos$^{1}$}
    \address{$^{1}$Northwestern University, Evanston, IL, USA \\
      $^{2}$University of California, Berkeley, CA, USA \\
      $^{3}$Eastern Switzerland University of Applied Sciences, Rapperswil, SG, CH}
}

\begin{document}
%
\maketitle
\begin{abstract}
Robustness has become one of the most critical problems in machine learning (ML). The science of interpreting ML models to understand their behavior and improve their robustness is referred to as explainable artificial intelligence (XAI). One of the state-of-the-art XAI methods for computer vision problems is to generate saliency maps. A saliency map highlights the pixel space of an image that excites the ML model the most. However, this property could be misleading if spurious and salient features are present in overlapping pixel spaces. In this paper, we propose a caption-based XAI method, which integrates a standalone model to be explained into the contrastive language-image pre-training (CLIP) model using a novel network surgery approach. The resulting caption-based XAI model identifies the dominant concept that contributes the most to the models prediction. This explanation minimizes the risk of the standalone model falling for a covariate shift and contributes significantly towards developing robust ML models. Our code is available at \url{https://github.com/patch0816/caption-driven-xai}.

\end{abstract}

\section{Introduction}
\label{sec:introduction}
\begin{figure}[htbp]
    \centering
    \includegraphics[width=0.75\columnwidth]{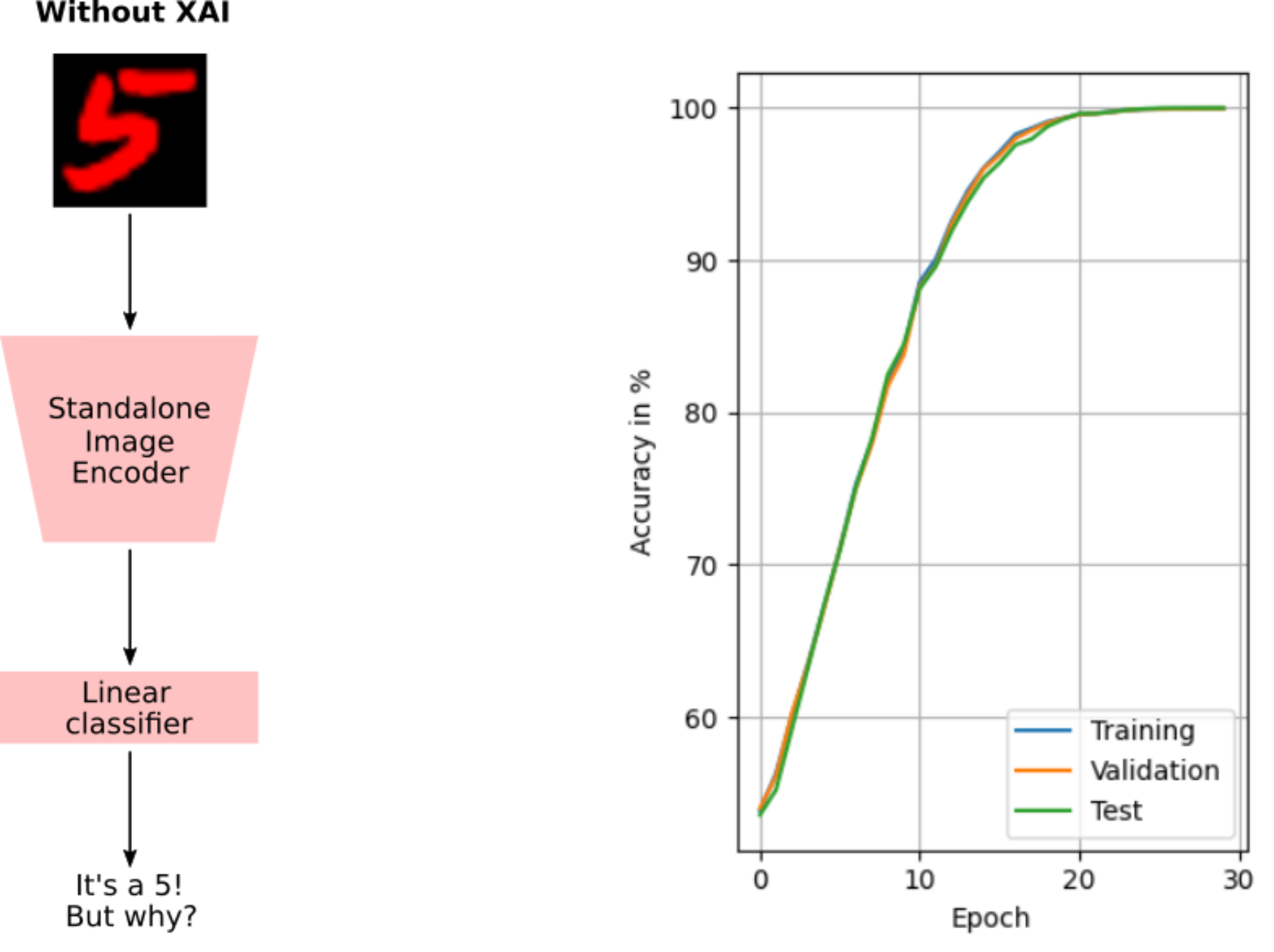}
    \caption{The standalone ResNet-50 model (Red) consists of an image encoder and a fully-connected linear classifier. The learning curves indicate a low bias, low variance ML model. Whether the ML model is biased without using XAI cannot be stated with certainty.
    }
    \label{fig:paper_1_situation}
\end{figure}

The fundamental property of ML models is that they are not explicitly programmed but learn from data instead. This attribute makes advanced ML models very powerful but challenging to interpret. With the ever-increasing capabilities and importance of ML models at the core of many applications, there is a need to prove their robustness, which represents one of the most crucial research areas in artificial intelligence (AI).\cite{adadi2018peeking} A robust ML model's performance in real-world situations deviates only marginally from the test performance, even if one or more features change drastically due to unforeseen circumstances. Expressing it differently, robustness refers to a model's ability to resist being fooled. In theory, the training, validation, and test datasets are sampled from the same data distribution. The temptation to deploy a low bias, low variance ML model as shown in Fig.~\ref{fig:paper_1_situation} is high. In a real-world scenario, there is always a risk involved that the data used for the training, validation, and test datasets does not accurately reflect the data distribution the deployed model faces. This distribution shift between the data used during the development of the model and the deployed model is designated as a covariate shift \cite{neuhaus2023spurious}. A covariate shift may be responsible for a model working in the lab for its intended task while failing in the real world. This characteristic is especially challenging in high-stakes environments, e.g., in medicine, where a patient could suffer from incorrect predictions made by an ML model \cite{degrave2021ai}. One evident approach to avoid a covariate shift is to ensure that the data for the development of the model reflects the real-world perfectly, but this is by no means a trivial task. Another approach is to use XAI methods. One of the state-of-the-art XAI methods to improve the robustness of ML models for computer vision problems is to generate saliency maps. There is a large variety of possibilities to generate saliency maps using class activation maps (CAM) \cite{zhou2016learning}, gradient-weighted CAM (Grad-CAM) \cite{selvaraju2017grad}, or learning important features CAM (LIFT-CAM) \cite{jung2021towards}, which estimates shapley values to weight the linear combination of activation maps by their marginal contribution to the explanation. All saliency map methods highlight the pixel space of an image that excites the model the most. \cite{Ahn_2024_CVPR} However, this property could be misleading if spurious and salient features are present in overlapping pixel spaces \cite{rudin2019stop}. The work of Bau et al. \cite{bau2019gandissect} about the GAN dissection method suggests that it is essential to understand the internal concepts of a model since these insights can help to improve the model's behavior. Their network dissection method \cite{bau2020units} demonstrates the generalizability of individual units responding to specific high-level concepts not directly represented in the training dataset. Measuring the alignment between the unit response and a set of concepts drawn from the broad and dense segmentation dataset \cite{oikarinen2023clipdissectautomaticdescriptionneuron} enables to define units as specific concept detectors. Inspired by the work on discovering concepts by Bau et al., our method, the caption-based XAI method, incorporates text to enhance the explanation. The main contribution of this paper solves the problem of identifying the dominant concept in multimodal units and therefore revealing a potential covariate shift before the deployment of the standalone model. Additionally, the caption-based XAI method works reliably even if spurious and salient features are present in overlapping pixel spaces. The demonstration of the caption-based XAI method in this paper uses a biased dataset, which leads to a biased standalone model. The biased dataset contains a covariate shift between the train, validation, and test datasets (representing the available data during the model development) and the real-world dataset (representing real-world data after deployment). The objective of the standalone model is to classify handwritten digits with the values five and eight from the MNIST dataset \cite{lecun1998mnist}. In the data available during the model development, all digits with the value five are colored red and all digits with the value eight are colored green. In the real-world dataset, the color assignments are random. This difference in the color assignment is responsible for the covariate shift. The caption-based XAI method aims to identify the dominant concept of the standalone model.

\section{Proposed method}
\label{sec:proposed_method}

\begin{figure}[htbp]
    \centering
    \includegraphics[width=\columnwidth]{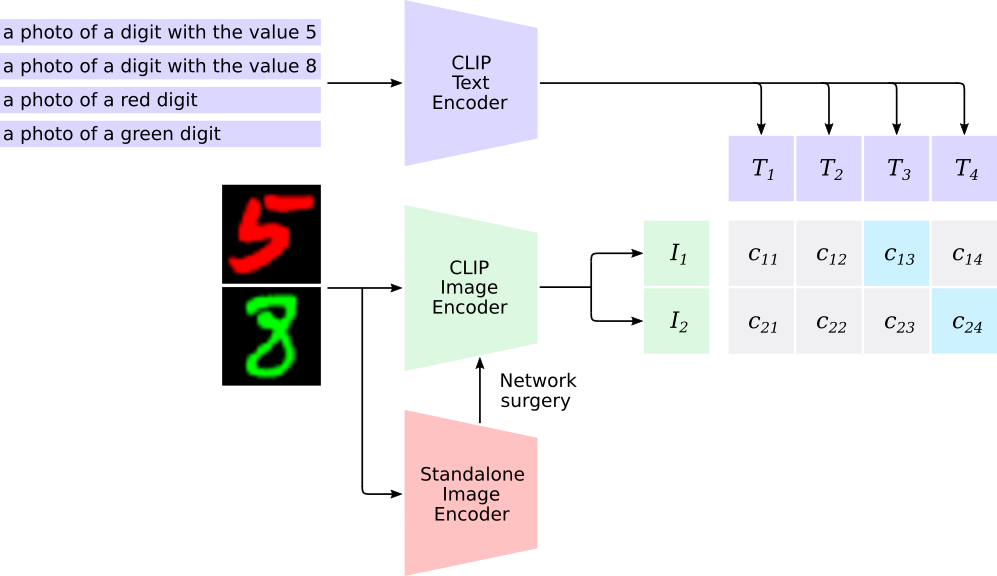}
    \caption{CLIP is the core component of the proposed caption-based XAI model. Using CLIP's text encoder (Purple) and image encoder (Green), the resulting embedding similarities reveal what CLIP's image encoder is focusing on by using the captions. The largest embedding similarities are highlighted (Blue). The network surgery process allows integration of any standalone model into CLIP, so CLIP can explain what the standalone image encoder (Red) focuses on.
    }
    \label{fig:paper_clip}
\end{figure}

The proposed caption-based XAI method uses a network surgery process to transfer the properties from the standalone model to be explained into CLIP \cite{radford2021learning} by swapping similar activation maps from the standalone image encoder to the CLIP image encoder resulting in the caption-based XAI model as shown in Fig.~\ref{fig:paper_clip}. Derived from the Euclidean dot product, the cosine similarity denotes the alignment of the two embeddings $\boldsymbol{I}_i$ and $\boldsymbol{T}_j$ in CLIP's space of concepts.

\begin{equation}
    c_{ij} = cos_{ij}(\theta) = \frac{\boldsymbol{I}_i \cdot \boldsymbol{T}_j}{\| \boldsymbol{I}_i \| \cdot \| \boldsymbol{T}_j \|}
    \label{eq:cosine_sim}
\end{equation}

Using suitable captions, the texts describing dominant concepts in the images result in significant embedding similarities. \cite{bhalla2024splice} If these high scores primarily arise for the color descriptions, then the standalone model is color biased. If these high scores primarily arise for the shape descriptions, then the standalone model is focused on the shapes.

\subsection{Architecture}
\label{sec:architecture}
CLIP is the core component of the caption-based XAI model. Many different configurations are available for CLIP's text and image encoder. Throughout this paper, the CLIP text encoder is a masked self-attention transformer \cite{vaswani2017attention, radford2019language} and the CLIP image encoder is OpenAI's modified and pre-trained \cite{zhang2019making} residual neural network-52 (ResNet-52) \cite{he2016deep} model. The main modifications of the ResNet model are the addition of two convolutional layers in the first stage and the replacement of the average pooling layer with an attention pooling layer. There are 51 convolutional layers, one fully connected layer and two pooling layers in the CLIP image encoder.

The standalone image encoder to be explained is a ResNet-50 model, which has been pre-trained on the ImageNet dataset \cite{imagenet_challenge} and finetuned for the MNIST binary classification task. There are 49 convolutional layers, one fully connected layer and two pooling layers in the standalone model. 

\begin{figure}[htbp]
    \centering
    \includegraphics[width=\columnwidth]{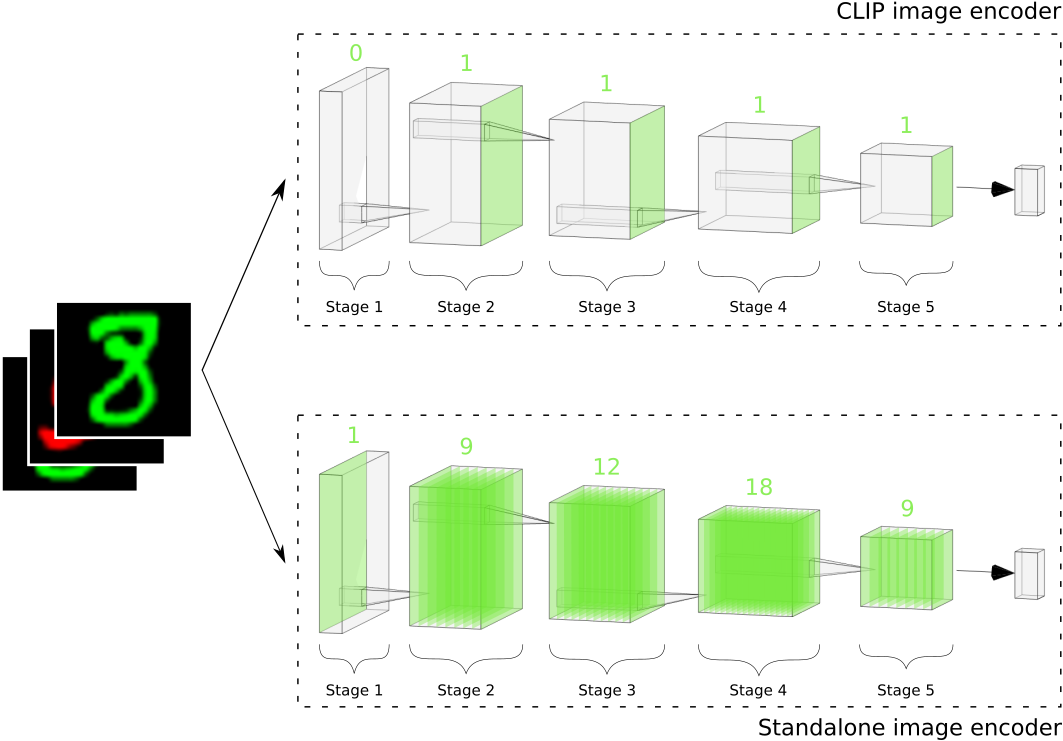}
    \caption{All activation maps within the four convolutional layers in the CLIP image encoder and 49 convolutional layers in the standalone model are available for swapping.
    }
    \label{fig:activation_matching}
\end{figure}

Incorporating the properties from the standalone model to be explained into the CLIP image encoder is a balancing act. On the one hand, we want to have all the standalone model's properties integrated into the CLIP image encoder to obtain the most significant explanation. On the other hand, the learned concept space of the CLIP embedding similarities needs to be maintained. To address this balancing act, all activation maps from the 49 convolutional layers of the standalone model are available for the selection process to be incorporated into the CLIP image encoder in order to transfer as much information as possible, as shown in Fig.~\ref{fig:activation_matching}. Each convolutional layer has a specific number of kernels resulting in a total number of 22720 activation maps in the standalone model. To maintain as much of the CLIP concept space as possible, only the last convolutional layers of stages 2, 3, 4, and 5 of the CLIP image encoder are available to be swapped. The first stage is an exception to the rule and remains untouched. The motivation is that the first stage captures very similar low-level concepts in both the standalone and CLIP models. Another motivation is that the CLIP captions typically describe high-level concepts rather than low-level ones. Only four out of the 51 convolutional layers are available for swapping. Each convolutional layer has a specific number of kernels resulting in a total number of 3840 activation maps in CLIP's image encoder to be swapped.

\subsection{Statistics}
\label{sec:stats}
Feeding the training dataset of images $\boldsymbol{x}$ into the standalone model $S(\boldsymbol{x})$ and the CLIP image encoder $C(\boldsymbol{x})$ and retaining all activation maps $\boldsymbol{A}^S_{i}$ and $\boldsymbol{A}^C_{j}$ for all images allows us to compute the statistics for each activation map. The mean and the standard deviation are suitable measures to describe the Gaussian distributions of the retained activation maps. The distribution of activations of the activation map $\boldsymbol{A}^S_{i}$ in the case of the standalone image encoder is described with the mean $\boldsymbol{\mu}^S_{i}$ and standard deviation $\boldsymbol{\sigma}^S_{i}$. The distribution of activations of the activation map $j$ in the case of the CLIP image encoder is described with the mean $\boldsymbol{\mu}^C_{j}$ and the standard deviation $\boldsymbol{\sigma}^C_{j}$.

\subsection{Activation matching}
\label{sec:activation_matching}
Due to the imbalance in the number of available activation maps between the standalone model to be explained and the CLIP image encoder, there is a need for a suitable selection process introduced as \emph{Activation Matching}. The objective is to find a subset of activation maps in the standalone model which are similar to the activation maps in the CLIP image encoder. Since the activation maps in ResNet models typically get smaller in size when moving to deeper layers due to pooling operations or the use of convolutional kernels with a stride of two, the activation maps need to be transformed into a comparable format of equal size and scale. In order to get activation maps of the same size, the smaller one of the two is upscaled using bilinear interpolation. The scales of the activation map $\boldsymbol{A}$ of standalone model $S$ and the CLIP image encoder $C$ are adjusted using a standard scaler and the model's respective statistics $\boldsymbol{\mu}$ and $\boldsymbol{\sigma}$.

\begin{equation}
    \boldsymbol{N}_{i} = \frac{\boldsymbol{A}_{i} - \boldsymbol{\mu}_{i}}{\boldsymbol{\sigma}_{i}}
\end{equation}

These scaled activation maps, $\boldsymbol{N}^S_{i}$ and $\boldsymbol{N}^C_{j}$, are used to compute the scores as a measure of similarity between each activation map of the standalone model and the CLIP image encoder. Correlation is used as the measure of similarity between two activation maps. Let $\boldsymbol{N}^S_{i}$ denote the scaled activation map of the standalone model $S$ within batch $B$ with width $W$ and height $H$. We apply a similar notation to $\boldsymbol{N}^C_{j}$. We then define the correlation coefficient $Z_{ij}$ between activation maps $\boldsymbol{N}^S_{i}$ and $\boldsymbol{N}^C_{j}$ as

\begin{equation}
    Z_{ij} = \frac{\sum_{b=1}^{B} \sum_{w=1}^{W} \sum_{h=1}^{H} \boldsymbol{N}^S_{biwh} \cdot \boldsymbol{N}^C_{bjwh}}{B \cdot W \cdot H}
\end{equation}\vspace{+3pt}

Each of the correlation coefficients $-1 \le Z_{ij} \le 1$ is used as an entry to form the activation matching score matrix $\boldsymbol{Z}$ of dimensionality $22720 \times 3840$, since $i=1, .., 22720$ and $j=1, .., 3840$.

\subsection{Network surgery}
\label{sec:network_surgery}

We now scan the activation matching score matrix to determine the largest entries indicating the activation maps that need to be swapped. Swapping two activation maps carries two challenges. The first challenge is the different scales of the two activation maps. The scaled activation map $\boldsymbol{N}^S_{i}$ to replace the CLIP image encoder activation map $\boldsymbol{N}^C_{j}$ is first scaled to form $\boldsymbol{A}^X_{j}$ according to

\begin{equation}
    \boldsymbol{A}^X_{j} = \boldsymbol{N}^S_{i} \cdot \boldsymbol{\sigma}^C_{j} + \boldsymbol{\mu}^C_{i}
\end{equation}

The second challenge is to upscale the activation map from the standalone model to the original size of the activation map from the CLIP image encoder using bilinear interpolation to integrate $\boldsymbol{A}^X_{j}$ perfectly between its neighboring layers.

\newpage
\section{Experiments}
\label{sec:experiments}

\begin{figure}[htbp]
    \centering
    \includegraphics[width=0.75\columnwidth]{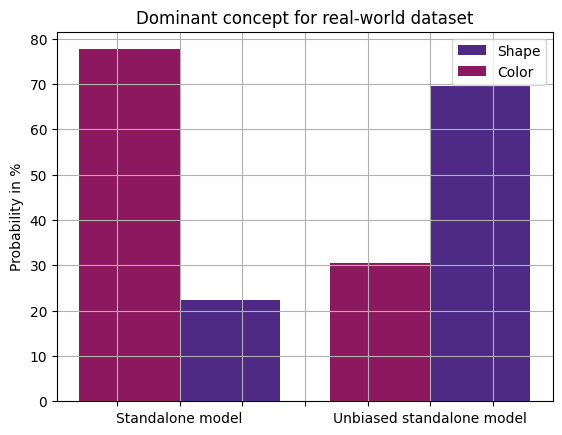}
    \caption{Color is the dominant concept for the standalone model and shape is the dominant concept for the unbiased standalone model.
    }
    \label{fig:dominant_concept_real_world_4}
\end{figure}

This section presents the results of the proposed caption-based XAI method applied to the standalone model using the colored MNIST test dataset of handwritten digits with the values five and eight. The objective is to identify the dominant concept of the standalone model. Given the four captions shown in Fig.~\ref{fig:paper_clip} during inference of the caption-based XAI model, the changes of the cosine similarities over the whole test dataset enable us to obtain statistically significant results. The difference in the cosine similarities before and after swapping is analyzed to exclude any initial bias from the CLIP model and to capture the influence of the network surgery exclusively. The number of correct/incorrect shape and color classifications can be aggregated by their common \emph{shape} or \emph{color} concept. The representation of the aggregated concepts reveals the dominant \emph{color} concept of the standalone model as shown in Fig.~\ref{fig:dominant_concept_real_world_4}, which is not apparent from Fig.~\ref{fig:paper_1_situation}.

In an ideal world with a perfect network surgery procedure, the probability for the concept \emph{color} should equal $100\%$ and $0\%$ for the concept \emph{shape} to identify the color bias. Due to the limitations of the network surgery approach, which incorporates $\frac{3840}{22720} = 16.9\%$ of all activation maps from the standalone model into the caption-based XAI model, the probabilities of the concepts \emph{shape} and \emph{color} need to be compared to each other.

\begin{figure}[htbp]
    \centering
    \includegraphics[width=0.75\columnwidth]{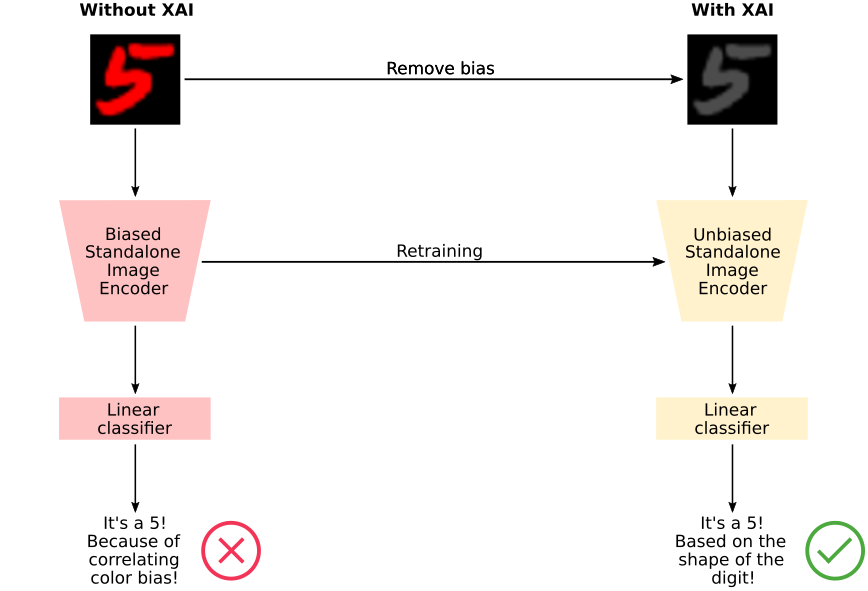}
    \caption{The caption-based XAI method identifies the color feature as the dominant feature. Removing the color feature and retraining makes the standalone model robust. The captions of the caption-based XAI model identify the shift from the color to the shape feature.
    }
    \label{fig:paper_3_xai}
\end{figure}

The caption-based explainable AI model successfully identifies a color bias in the standalone model as demonstrated in Fig.~\ref{fig:dominant_concept_real_world_4}. This explanation can be used to de-bias the dataset using a color-to-grayscale pre-processor and train a new unbiased standalone model as shown in Fig.~\ref{fig:paper_3_xai}.

Incorporating the grayscale unbiased standalone model into the caption-based explainable AI model using network surgery results in a counterintuitive effect. Due to the four shape-focused and color-focused captions, the caption-based explainable AI model can still predict a red or green digit. Part of the reason for this behavior is that the grayscale images are still red, green and blue color images but with the same values on all three channels. Since there are no colored digits in the grayscale dataset anymore, the \emph{correct color} and \emph{incorrect color} numbers aggregate to \emph{any color}, which should be equal to zero in an ideal world, but CLIP is not perfect. Incorporating the unbiased standalone model into the caption-based explainable AI model using the network surgery procedure identifies the concept \emph{shape} to be the dominant concept and confirms the removal of the color bias as shown in Fig.~\ref{fig:dominant_concept_real_world_4}. Visualizing the aggregated measures by their respective \emph{shape}/\emph{color} concepts results in a significant shift of the dominant concept from \emph{color} in the standalone model to \emph{shape} in the unbiased standalone model, as shown in Fig.~\ref{fig:dominant_concept_real_world_4}.

\section{Conclusion}
\label{sec:conclusion}
This work introduces a new approach called the caption-based XAI method to explain convolutional neural networks. Using a novel network surgery method, a standalone model to be explained is incorporated into CLIP. The resulting caption-based XAI model successfully identifies the dominant concept that contributes the most to the model's prediction. This finding enables us to improve the standalone model and increase its robustness accordingly before deploying it into the real-world. This property could be especially insightful in medical applications to confirm or debunk doctor's preconceived notions. The most promising result is the superiority of the novel XAI method over saliency maps in situations where spurious and salient features are present in overlapping pixel spaces. The central thesis validated by this work is that a deeper understanding of the dominant concepts in convolutional neural networks is fundamental and can be used to improve the model's robustness. Our findings suggest that this novel XAI method should not just be seen as a pure debugging tool but as a necessary prerequisite before deploying any machine vision convolutional neural network model.

\vfill\pagebreak



\bibliographystyle{IEEEbib}
\bibliography{strings,refs}

\end{document}